\pdfoutput=1
\documentclass[11pt]{article}

\usepackage[final]{latex/acl}

\usepackage{times}
\usepackage{latexsym}

\usepackage[T1]{fontenc}
\usepackage[utf8]{inputenc}

\usepackage{microtype}

\usepackage{inconsolata}

\usepackage{float}
\usepackage{graphicx} % Required for inserting images

\usepackage{amsmath}
\usepackage{amssymb}
\usepackage{caption}
\usepackage{xcolor}

\usepackage{xurl} % Handles URL wrapping
\usepackage{breakurl} % Backup for traditional URLs
\usepackage{hyperref} % Hyperlinks and URL management

\usepackage{booktabs} %for using specialrule
\usepackage{multirow} %for using multirow cline
\usepackage{multicol}
\usepackage{amsfonts} %for mathbb
\usepackage{natbib} % for using \cite{}
\usepackage{makecell}
\usepackage{array}
\newcolumntype{H}{>{\setbox0=\hbox\bgroup}c<{\egroup}@{}}\newcolumntype{C}[1]{>{\centering\arraybackslash}p{#1}}

\usepackage{arydshln} % always keep {array} before it
\usepackage{listings} % inclusion of algorithm/code
\usepackage{fancyvrb}
\DefineVerbatimEnvironment{CustomVerbatim}{Verbatim}{commandchars=\\\{\}}
\usepackage{subcaption}

\usepackage{printlen} % Include the package for printing textwidth in cm

\usepackage{algorithm}
\usepackage[noend]{algpseudocode}

\newcommand{\1}{\tiny $^\triangledown$}
\newcommand{\2}{\tiny $^\vartriangle$}
\newcommand{\3}{\tiny $^\blacktriangle$}
\newcommand{\4}{$^*$}

\usepackage[autostyle]{csquotes}
\usepackage{enumitem}

\title{Robust Bias Detection in MLMs and its Application to Human Trait Ratings}

\author{Ingroj Shrestha \\ 
        University of Iowa \\
    \texttt{ingroj-shrestha}\\ \texttt{@uiowa.edu} \\
  \And
    Louis Tay \\
    Purdue University \\
    \texttt{stay@purdue.edu} \\
    \And
  Padmini Srinivasan \\
  University of Iowa \\ 
  \texttt{padmini-srinivasan}\\ \texttt{@uiowa.edu}\\
  }

\begin{document}
\maketitle
\begin{abstract}
\noindent There has been significant prior work using templates to study bias against demographic attributes in MLMs. However, these have limitations: they overlook random variability of templates and target concepts analyzed, assume equality amongst templates, and overlook bias quantification.
Addressing these, we propose a systematic statistical approach\footnote{Our code and data are available at \url{https://github.com/IngrojShrestha/robust_mlm_bias_detection_human_trait_ratings}}
to assess bias in MLMs, using mixed models to account for random effects, pseudo-perplexity weights for sentences derived from templates and quantify bias using statistical effect sizes.
Replicating  prior studies, we match on bias scores in magnitude and direction with small to medium effect sizes.
Next, we explore the novel problem of gender bias in the context of  \textit{personality} and \textit{character} traits, across seven MLMs (base and large).
We find that MLMs vary; ALBERT is unbiased for binary gender but the most biased for non-binary \textit{neo}, while  RoBERTa-large is the most biased for binary gender but shows small to no bias for \textit{neo}.
There is some alignment of MLM bias and findings in psychology (human perspective) - in \textit{agreeableness} with RoBERTa-large and \textit{emotional stability} with BERT-large. 
There is general agreement for the remaining 3 personality dimensions: both sides observe at most small differences across gender.
For character traits, human studies on gender bias are limited thus comparisons are not feasible.

\end{abstract}

\section{Introduction}

\noindent Pre-trained Masked Language Models (MLMs) (e.g., BERT \cite{devlin-etal-2019-bert}) are valuable but show systematic biases favoring certain demographics.
E.g., these indicate men as more likely to be engineers, and women as being more emotional \citep{gallegos2023bias,lee2018detecting, parikh2019addressing, booth2021integrating}.
These biases amplify societal marginalization and discrimination in automated decision-making and diminish trust in AI systems \cite{solaiman2023evaluating}.
Our research contributes to the active stream on bias detection in MLMs.  
Our \textbf{first goal} is to propose methods correcting  limitations commonly exhibited by MLM  bias detection approaches that hinder robust inferences.
Our \textbf{second goal} is to use our methods to gauge if MLMs are biased across gender in the novel domain of perceptions of character and personality.
Such bias would undoubtedly make it risky to use MLMs in socially critical contexts such as hiring and promotion decisions.

\subsection{Methodological limitations}

MLM bias detection typically begins with sentence templates from which parallel sets of sentences are derived, one for each demographic group considered. 
In essence, templates specify a universe of sentences used to `probe' an MLM to gauge how it associates a demographic group and a concept representing a domain such as \emph{employment}; this assessment yields a score.
Templates are essential since it is infeasible to consider all possible relevant sentences related to a target concept.  

With few exceptions, template selection and the derivation of probe sentences are done manually. 
Template sets also tend to be inherited from one paper to the next.
A key problem is that it is typical to treat all templates used as equal.
However, there could be variations across templates that cloud the detection of MLM bias.
We handle this problem using a \emph{mixed effects model} where template variations are handled as random effects.  
Using parallel logic, we also handle variations across different domain words as random effects (e.g., variations across words for different jobs).
Again, such variations should not confound bias assessment.

A second problem is that probe sentences derived from the same template can vary greatly.
For example, both \textit{She is a considerate person} and \textit{She is a concerned person} derive from the template \textit{[gendered-word] is a/an [trait-word] person}.
They have pseudo-perplexity scores of 1.8 and 13.8, respectively, when evaluated using BERT-large (uncased version).
Given this difference treating these as equivalent for bias detection is also risky. 
It makes sense to weigh sentence bias estimate by commonality (estimated as pseudo-perplexity).

The next problem is one of making statistically robust inferences. The minimal approach is to see if the association score difference across demographic groups is statistically significant or not. 
However, in some cases, even this is absent, relying only on the raw difference in association scores to assess bias \cite{limisiewicz-marecek-2022-dont,guo2022auto,kaneko-bollegala-2021-debiasing}.
We consider it important to go beyond significance and consider \emph{effect size}.
Effect size tells us how much of an observed difference in association scores is explained by the demographic variable of interest.
There may be a sizable and significant difference, but if the effect size is small, then the bias is also small.  
This can happen if other factors unaccounted for in the study are responsible for score differences.
Our first goal is to advocate for a bias detection methodology that does not have these limitations.

\subsection{Character and personality perceptions}

\noindent MLMs have become deeply entrenched in different societal contexts.  In psychology, MLMs are being used to estimate human attributes like personality and character from social media texts \cite{park2015automatic, liou2023online, pang2020language}. 
In organizational settings, they are used in hiring to infer individual attributes from language in job applications \cite{thompson2023deep}, interviews \cite{hickman2022automated}, surveys \cite{speer2023turning}, video interviews and resumes \cite{booth2021bias, gagandeep2023evaluating}.
Clearly, biases in these MLM applications would jeopardize the integrity of outcomes and perpetuate stereotypes.
While several studies in psychology study bias (or at least differences) when humans rate males and females at least on personality traits, MLMs have not been assessed in the same context.
Thus, our second goal is to assess MLMs for gender biases in character and personality ratings. 

We draw on two major psychological frameworks specifying key human traits. One known as \textit{human virtue} or \textit{character traits}
% character strengths or 
\cite{peterson2004character} specifies key positive traits of individuals. Another specifies \textit{personality traits}, which are enduring descriptive characteristics of individuals. 
We use lexical approaches based on adjectives describing people as outlined by \citet{john1988lexical}.
From lexical studies, there are four key character dimensions (\textit{empathy}, \textit{order}, \textit{resourceful}, \textit{serenity}) \cite{cawley2000virtues} and five personality dimensions (\textit{extroversion}, \textit{agreeableness}, \textit{conscientiousness}, \textit{emotional stability}, \textit{openness}) \cite{goldberg1992development}.
Our second goal is to assess MLMs for gender biases along these nine trait dimensions.

\noindent In summary:
\begin{enumerate}[noitemsep, nolistsep]
    \item We propose a better bias detection methodology for MLMs achieved with a mixed effect model accommodating fixed and random effects.  We also weigh probe sentences and estimate effect size.

    \item We assess seven MLMs for gender bias in the novel domain of character and personality traits. Gender bias detection is critical for the societal contexts in which MLMs are used.

\end{enumerate}

We first describe our bias detection method. Then we present results from two replication studies followed by our main results on MLM bias in human trait perception with additional analysis. We then present related works and conclusions ending with limitations and an ethics statement.

\section{Methodology}\label{measuring-association}

\noindent We follow the standard template-based approach to estimate bias in MLMs \cite{gallegos2023bias,delobelle2022measuring,stanczak2021survey}.
A template is a sentence structure with two variables representing a \textit{demographic} attribute word ($A$) and a \textit{domain} target word ($T$), along with other words. 
Here, the attribute is gender, and the target is human traits (character/personality). 
Templates are used to derive probe sentences ($S_1$, $S_2$,..., $S_n$).
Bias is assessed by analyzing the MLM estimates of the association between attribute and target words in probe sentences.  

\vspace{0.5em}
\noindent\textbf{Measuring association:} We follow the  approach of \citealp{kurita-etal-2019-measuring}. 
Briefly, we mask the attribute $A$ in a probe sentence ($S_{\text{masked}}^{(A)}$), provide it as input to the MLM, and obtain the likelihood\footnote{While recognizing that this is actually a pseudo-likelihood \cite{salazar-etal-2020-masked}, we use the common approach of using it as a proxy for likelihood.} of the attribute ($p_{A}$), i.e., 
$p_{A} = p_{\text{MLM}}(\text{[MASK]}=A|S_{\text{masked}}^{(A)};\theta$), where $\theta$ represents the MLM's parameters.
However, since the likelihood of predicting different attribute values could differ even in the absence of a target, we also compute the `implicit prior bias' across attribute values. 
To do this, we mask both attribute and target and then obtain the likelihood of the attribute value ($S_{\text{masked}}^{(A,T)}$).
We refer to this as $p_{\text{prior}} = p_{\text{MLM}}(\text{[MASK]}=A|S_{\text{masked}}^{(A,T)};\theta)$.
Association score ($\text{association}_{\text{score}}$) is $\log \left( \frac{p_{A}}{p_{\text{prior}}} \right)$.
Where the MLM splits attribute word into multiple tokens, we take the product of the likelihood of sub-tokens, as commonly practiced \cite{
% fadeeva-etal-2024-fact,
shahriar-barbosa-2024-improving,ahn-oh-2021-mitigating}.

\noindent \textbf{Masking example:} 

\noindent $S_{i}$: "The lady is known for her empathy."

\noindent $S_{i,\text{masked}}^{(A)}:$  "The [MASK] is known for [MASK] empathy." 

\noindent $S_{i,\text{masked}}^{(A,T)}:$ "The [MASK] is known for [MASK] [MASK]."

\noindent Note that we also mask gendered pronouns (e.g., `her') to prevent leakage of gender information. For such cases, while computing $p_{A}$ and $p_{\text{prior}}$, we consider the likelihood of the attribute word in the first [MASK] position only (e.g. see dataset description of \citet{bartl-etal-2020-unmasking} paper).

\subsection{Templates}\label{template-sentences}

\noindent Templates provide the skeletal structure for probe sentences. Clearly, one can only consider a sample of all possible templates 
% kurita-etal-2019-measuring
\cite{limisiewicz-marecek-2022-dont,ahn-oh-2021-mitigating,bartl-etal-2020-unmasking}.
The dominant approach has been manual template design \cite{doughman2023fairgauge,felkner-etal-2023-winoqueer,mei2023bias,delobelle2022measuring}
%gallegos2023bias
%
with a few exceptions such as  \citet{guo2022auto,shin2020autoprompt,liang-etal-2020-towards}.

We select templates using a semi-automatic process designed to capture common expressions of human traits using Wikipedia\footnote{Our approach is somewhat analogous to research on bias in auto-regressive language models where Wikipedia sentences are truncated semi-automatically to create prompts \cite{lucy2021gender,dhamala2021bold}.} and GPT-4
(see Appendix \ref{template-selection-overview} for an overview). 
We have two template types. 
\textit{Direct} explicitly include the word \textit{personality}, and \textit{Indirect} do not.
The idea is to see if the word \textit{personality} guides the model more effectively. 
E.g., \textit{clean} may then be more easily perceived as representing a personality trait instead of its more common meaning of physical cleanliness.

Our 6 templates (Table \ref{tab:templates}) align with the common practice of using 2-5 templates 
% kurita-etal-2019-measuring
\cite{steed2022upstream,limisiewicz-marecek-2022-dont,bartl-etal-2020-unmasking,qian2019reducing}.  
However, unlike previous research, we do not assume that all templates are equal for bias detection.

\vspace{0.5em}
\noindent\textbf{Attributes and targets:} Attributes are 94 pairs of gender-denoting words adapted from \citet{kaneko-bollegala-2021-debiasing} and
listed in Table \ref{tab:attributes}.
Targets are character trait words \cite{cawley2000virtues} listed in Table \ref{tab:targets_virtue} and personality trait words \cite{goldberg1992development} listed in Table \ref{tab:targets_big_five}. Tables are in the Appendix \ref{attribute-values} and \ref{target-values}.
Again, we handle variations across target words as random effects in our model.

\vspace{0.5em}
\noindent\textbf{Sentence generation from templates:}  We also pay particular attention to selecting the appropriate article (limited to \textit{a/an}) and determiner (\textit{the}) or pronoun (limited to \textit{my, your, our, their}) to form wholesome sentences. 
To avoid arbitrariness, we do this by estimating the psuedo-perplexity of a sentence. Using the MLM for which we are conducting bias detection, we select the candidate (e.g., pronoun) with the least perplexity for each gender.
When selections differ across genders, e.g., \textit{my father} and \textit{your mother}, we add the alternatives \textit{your father} and \textit{my mother} for balance.

Across templates, the average sentence count ranges from 1,447 to 4,119 for character traits and 1,437 to 1,757 for personality traits, with a small coefficient of variation 
of 3.5\% to 10.8\%. Since sentence selection is MLM specific, the numbers can vary within each template and trait dimension.  Thus, we provide averages to ensure a consistent overview in Table \ref{tab:num_sentences_dist} in Appendix.

\begin{table*}[htbp]
    \scriptsize
    \centering
    \resizebox{\textwidth}{!}{
    \begin{tabular}{p{0.8cm}p{0.3cm}p{12.3cm}@{}}
    \toprule
    \textbf{Category} & \textbf{TID} & \textbf{Templates} \\
    \midrule
    \multirow{4}{*}{Indirect}    & $\text{t}_1$    & {[}DET/PRONOUN{]} {[}attribute{]} is {[}ARTICLE{]} {[}target{]} person. e.g. my father is a cautious person. \\
                          \cmidrule{2-3}  & $\text{t}_2$     & {[}DET/PRONOUN{]} {[}attribute{]} is {[}target{]}. e.g. my mother is cautious.\\
                                
    \midrule
    \multirow{10}{*}{Direct}       & $\text{t}_3$     & {[}DET/PRONOUN{]} {[}attribute{]} possesses {[}ARTICLE{]} {[}target{]} personality. \\ 
    & & e.g. my father possesses a cautious personality.   \\

                                \cmidrule{2-3}  & $\text{t}_4$      & {[}DET/PRONOUN{]} {[}attribute{]} is known for {[}PRONOUN{]} {[}target{]} personality. \\ & & e.g. she is known for her cautious personality. \\
                                   
                           \cmidrule{2-3}
                                & $\text{t}_5$       & People admire {[}DET/PRONOUN{]} {[}attribute{]} because of {[}PRONOUN{]} {[}target{]} personality. \\
                                &    & e.g. people admire him because of his cautious personality. \\

                        \cmidrule{2-3}        & $\text{t}_6$       & {[}DET/PRONOUN{]} {[}attribute{]}'s {[}target{]} personality is valued at {[}PRONOUN{]} work. \\ 
                         &       & e.g. the woman's cautious personality is valued at her work.\\
                               
    \bottomrule                                   
    \end{tabular}
    }
    \captionsetup{justification=centering}
    \caption{
    \footnotesize
    Templates (TID: template id, attribute: gendered-words, target: character trait words/personality trait words (above examples use character trait words), Determiner (DET): the,  PRONOUN: my, your, our, their)}
    \label{tab:templates}
\vspace{-4mm}
\end{table*}
\raggedbottom

\subsection{Linear mixed model configuration} \label{model-configuration}

\noindent 
In our mixed effect model \cite{baayen2008mixed}, represented below, \textit{gender} (values: \textit{male} or \textit{female}) is a fixed effect (predictor) and association score ($\text{association}_{\text{score}}$) the response variable.
Unlike prior research, we account for  variability across  templates and trait words
as random effects represented using standard notation (1|random\_effect).
These make the statistical estimates more generalizable, which is a critical feature of our methodology. We use the lme4 \cite{lme4} package in R to fit the mixed models.
This package incorporates and estimates the influence of both fixed and random effects in a statistically robust manner.

Besides structure, sentences derived from templates can also differ in their popularity. Thus, we weigh each sentence using pseudo-perplexity \cite{salazar-etal-2020-masked}.
Specifically, we give higher weights to sentences with lower pseudo-perplexity calculated using the same MLMs being analyzed for bias.
This weight is introduced during model fitting, adjusting residual variance rather than directly modifying the association score. The overall linear mixed-effect model is as follows:

\vspace{0.5em}

\noindent $\text{model}_{\text{lme}}$: $\text{association}_{\text{score}}$ $\sim$ gender $+$ (1 $|$ template) + (1 $|$ trait\_words)

\vspace{0.2em}
where, weight = 1 / (sentence pseudo-perplexity)

\vspace{0.5em}
Pseudo-perplexity is computed by masking one word at a time in the sentence and obtaining the likelihood of the original word. It is the exponential of sum of logs of losses in predicting original words. Following common notation we refer to this as (an estimate of) perplexity. 
 
\subsection{Bias assessment}\label{bias-assessment}

\textbf{Bias score:} This score is given by the coefficient of the \textit{gender} variable in the model. It represents the difference in association scores between genders across targets and templates.
A positive (negative) bias score refers to bias against females (males).

We make robust conclusions as follows.
First, we test significance of bias score (95\%, using Welch's t-test \cite{welch1947generalization}). While somewhat common practice  
(e.g., 
% kurita-etal-2019-measuring
\citet{koksal2023language, bartl-etal-2020-unmasking}), some bias papers do not test significance (e.g., 
% limisiewicz-marecek-2022-dont,ousidhoum2021probing,
\citet{guo2022auto,kaneko2022unmasking,ahn-oh-2021-mitigating}).
We also consider effect size, a step rarely taken in the bias literature (e.g., while \citet{dayanik2022analysis,bartl-etal-2020-unmasking,kurita-etal-2019-measuring} measure effect size, \citet{kim2023race,guo2022auto,kaneko2022unmasking,limisiewicz-marecek-2022-dont} do not).
% \cite{sullivan2012using}. 
Effect size measures the magnitude of differences \emph{while accounting for variability within each gender group}.
In contrast raw score differences disregard within-group variability. Thus, we consider bias score, its significance \emph{and} effect size.

\vspace{0.5em} 
\noindent \textbf{Effect size measurement:} 
We choose $R^2$ over  Pearson correlation as our measure because $R^2$ accounts for relationships involving random effects, unlike Pearson correlation \textit{r}.
Since our main goal is to analyze bias across gender, we focus on $R^2$ for the \textit{gender} attribute only.
Assuming significance at 95\%, the higher the $R^2$ the more important gender is in explaining differences in association scores.
We follow $R^2$ interpretation guidelines provided by \citet{cohen1988statistical}: very small: [0, 0.01), small: [0.01, 0.09), medium: [0.09, 0.25), large: [0.25, 0.64), very large: [0.64, 1.0].
To understand the relative magnitude of small effect size, i.e., whether it is closer to medium or very small, we further break down it into three groups {\scriptsize $\triangledown$}: [0.01, 0.03),
{\scriptsize $\vartriangle$}: [0.03, 0.06),
and {\scriptsize $\blacktriangle$}: [0.06, 0.09).
We annotate medium to very large effect as $*$.

\vspace{0.5em}
\noindent \textbf{$R^2$ confidence intervals (CI):} %
We conduct 1000 
parametric bootstrap iterations.
In each,  
we sample with replacement to create a new dataset of the same size. 
The resulting 1000 $ R^2$ values are used to estimate the confidence intervals using the partR2 library \cite{stoffel2021partr2}. 

\vspace{0.5em}
\noindent \textbf{Determination of Bias:} As is standard, bias scores that are not significant indicate \textit{neutral} or \textit{unbiased} stance. 
In addition, we consider significant bias scores but with effect size, $R^2 < 0.01$, as \textit{unbiased}.

\subsection{MLMs assessed for bias}
We analyze four pre-trained MLMs, both base and large, except for distilbert:
bert-base-uncased (bert-large-uncased) \cite{devlin-etal-2019-bert}, 
roberta-base (roberta-large) \cite{roberta-paper}, 
albert-base-v2 (albert-large-v2) \cite{albert-paper}, 
distilbert-base-uncased \cite{Sanh2019DistilBERTAD}.
All models used are the uncased versions\footnote{All models except RoBERTa use WordPiece tokenization; thus we use the uncased version. RoBERTa uses Byte-Pair Encoding, supporting both cased and uncased text inherently.}. Note that both ALBERT models are uncased by default. We provide input text in lower case for all models for consistency. We implement MLMs using Hugging Face on NVIDIA Tesla P100 PCIE (16GB) GPU. Each MLM model took about 3 hours on average per trait.

\section{Results}

\subsection{Prior work replication results}\label{prior-work-replication}

\noindent First, we replicate our methods on two prior studies \cite{bartl-etal-2020-unmasking,limisiewicz-marecek-2022-dont}, 
that share key methodological features, namely the use of templates, the same association scores to measure bias (Section \ref{measuring-association}), and focus on gender bias. 
We explore their targets:  \textit{profession} in \citet{bartl-etal-2020-unmasking} and \textit{profession/non-profession} indicated by \textit{Nouns} in \citet{limisiewicz-marecek-2022-dont}\footnote{A third work by \citet{kurita-etal-2019-measuring}, 
also shares the same methodological features.
However, we do not replicate it given its significantly small dataset size.}.
We aim to see if we obtain comparable results despite using our analytic methods. 

\subsubsection{\texorpdfstring{\citet{bartl-etal-2020-unmasking}}{}}

\noindent \textbf{Overview:} The authors study gender bias w.r.t. profession for both English and German. 
We focus on their results for English.
Consistent with their work, we use bert-base-uncased MLM and their templates (their Table 1 \textit{English}).  
They consider 3 categories with 20 professions in each: \textit{Balanced}, \textit{Female} and \textit{Male}. 
\textit{Female} (\textit{Male}) refers to professions where females (males) dominate in the real world while in \textit{Balanced} professions both have roughly equal participation, decided using \citet{us2020employed}.
Consistent with their work, we compute averages of association scores (Section \ref{measuring-association}) for each gender across all templates (they call this \textit{Pre}) and then take bias score as male minus female average (Table 4 in their paper), i.e., \textit{m-f}.
In our mixed model approach (Section \ref{model-configuration}), professional words from their paper substitute for trait words.
While they consider 
the significance of bias scores and effect size, it is unclear whether the reported effect size relates to Pre-association or Post-association (after fine-tuning). So, we exclude their effect size in our comparison.

\begingroup
\begin{table}[htbp]
    \scriptsize
  \centering
    \begin{tabular}{@{}llcc@{}}
    \toprule
          &  \textbf{Method}     & \multicolumn{1}{p{5.5em}}{\textbf{Bias score}} & \multicolumn{1}{l}{\textbf{Effect Size ($R^2$)}} \\
    \midrule
    \multirow{2}[0]{*}{Balanced} & Prior work & 0.40  &   \\

         & $\text{model}_{\text{lme}}$ & 0.41  & 0.13\4 \\
    \cmidrule{2-4}
    \multirow{2}[0]{*}{Female} & Prior work & -1.18 &   \\

          & $\text{model}_{\text{lme}}$ & -0.83 & 0.24\4 \\
    \cmidrule{2-4}
    \multirow{2}[0]{*}{Male} & Prior work & 0.99  &   \\

          & $\text{model}_{\text{lme}}$ & 1.00  & 0.50\4 \\
    \midrule
    \multirow{2}[0]{*}{\begin{tabular}[l]{@{}l@{}}104\\\textit{Nouns}\end{tabular}}& Prior work & 0.35  &   \\
          
          & $\text{model}_{\text{lme}}$ & 0.40  & 0.06\3 \\
    \bottomrule
    \end{tabular}%
\caption{{
\footnotesize
Prior work replication results. 
The first six rows refer to \citet{bartl-etal-2020-unmasking}. Last two rows refer to \citet{limisiewicz-marecek-2022-dont}. All bias scores and effect sizes ($R^2$) are significant at 95\% confidence level. Please see Table \ref{tab:model2-results} legend for notation. }
}
\label{tab:replication_results}%
\vspace{-5mm}
\end{table}%
\raggedbottom
\endgroup

\vspace{0.5em}
\noindent \textbf{Results and Analysis:}Replication results are in the first 6 rows of Table \ref{tab:replication_results}.
Our bias scores are significant throughout. 
Our model, $\text{model}_{\text{lme}}$, scores are close to prior results in magnitude and the same in direction.
As before for \textit{Female} (\textit{Male}) professions bias is high and against males (females) while for the \textit{Balanced} professions bias scores of around 0.41 are lowest but still favoring males.

While our bias scores and direction match those of \citet{bartl-etal-2020-unmasking}, we provide additional meaningful analysis based on effect sizes.
Our model yields larger effect sizes -- medium to large 
: 0.13 (\textit{Balanced}), 0.24 (\textit{Female}) and 0.5 (\textit{Male}). 
E.g., a sizable 13\% of score variations in \textit{Balanced} are explained by gender differences.

\subsubsection{\texorpdfstring{\citet{limisiewicz-marecek-2022-dont}}{}}

\noindent \textbf{Overview:}
We focus on their investigation of gender bias in the context of 104 gender neutral professional and non-professional \textit{Nouns} (e.g., professional: `chef', `programmer', `painter'; non professional: `victim', `customer', `patient').
We use their templates (their Table 1), bert-base-cased MLM, and 
their masking strategy involving determiners, pronouns and the nouns of interest and compute bias score as in their paper.
We replace trait words with their \textit{Noun} words in our models. 
Notably, the authors do not consider bias score significance, effect size, or control for random effects.

\vspace{0.5em} 
\noindent \textbf{Results and Analysis:} As seen in the last two rows of Table \ref{tab:replication_results} 
our bias score is close to theirs (0.352, see column MEAN, row 1 in their Table 3) in magnitude and matches direction favoring males.
However, while it is statistically significant (95\% confidence), our effect size is small, 0.06. Thus, we conclude that the bias found in
\citet{limisiewicz-marecek-2022-dont}
is small.
This underlines the importance of considering random effects and of incorporating sentence weights.

\begin{table*}[htbp]
    \scriptsize
  \centering
  \setlength{\tabcolsep}{1.5pt} % adjust spacing between columns
    \begin{tabular}{p{1.5cm}p{2.6cm}ccccccc}
    \toprule
    \multicolumn{1}{l}{} & \textbf{Traits} & \multicolumn{7}{c}{\textbf{Language Models}} \\
    \midrule
          & \multicolumn{1}{r}{} & \multicolumn{4}{c}{\textbf{Base}} & \multicolumn{3}{c}{\textbf{Large}} \\
    \cmidrule{3-9}
          & \multicolumn{1}{r}{} & \textbf{BERT} & \textbf{RoBERTa} & \textbf{ALBERT} & \textbf{DistilBERT} & \textbf{BERT} & \textbf{RoBERTa} & \textbf{ALBERT} \\
    \midrule
    \multirow{4}[0]{*}{\begin{tabular}[l]{@{}l@{}}\textbf{Character}\\\textbf{traits}\end{tabular}} & \textit{empathy} &-0.36\1 & -0.19 & -0.19 & -0.37\2 & 0.99\2  & -0.70\3 & -0.30 \\
          & \textit{order} & \textcolor{blue}{-0.08}  & 0.12  & \textcolor{red}{-0.06} & -0.07 & 0.69\1  & -0.30\1 & -0.41 \\
          & \textit{resourceful} & -0.30 & -0.20 & \textcolor{blue}{-0.15}  & -0.22\1 & 0.86\2  & -0.74\2 & -0.24 \\
          & \textit{serenity} & -0.33 & -0.43\1 & -0.47 & -0.36\2 & 0.47\1  & -1.08\4 & -0.35 \\
        \cmidrule{2-9}
          
          & Effect Size ($R^2$) & [1E-3, 1.03E-2] & [1E-3, 1E-2] & [0, 2E-3] & [1E-3, 3.7E-2] & [1E-2, 3.5E-2] & [1E-2, 0.127] & [1E-3, 3E-3]  \\
    \midrule
    \multirow{5}[0]{*}{\begin{tabular}[l]{@{}l@{}}\textbf{Personality}\\\textbf{traits}\end{tabular}} & \textit{extroversion} & -0.28 & -0.39\1 & -0.26 & -0.25\1 & 0.71\1  & -0.86\4 & -0.38 \\
          & \textit{agreeableness} & -0.16 & -0.21 & \textcolor{red}{-0.16} & \textcolor{red}{-0.05} & 0.61\1  & -0.77\3 & -0.36 \\
          & \textit{conscientiousness} & -0.20 & 0.54\1  & \textcolor{red}{0.05}  & -0.23\1 & 0.64\1  & -0.77\3 & -0.23 \\
          & \textit{emotional stability} & -0.18 & \textcolor{red}{-0.08} & -0.21 & -0.21\1 & 1.01\2  & -0.26 & -0.56 \\
          & \textit{openness} & -0.27 & 0.15  & -0.20 & -0.16 & 0.35  & -0.61\2 & \textcolor{red}{0.12} \\
          \cmidrule{2-9}      
    & Effect Size ($R^2$) & [2E-3, 4E-3] & [0, 2E-2] & [0, 1E-3]
    & [1E-3, 2E-2] & [9E-3, 4.1E-2] & [5E-3, 0.104] & [1E-3, 4E-3] \\
    \bottomrule
    \end{tabular}    
     \caption{
     \footnotesize
     Results for Seven MLMs using $\text{model}_{\text{lme}}$. 
    Values presented in each cell (e.g., -0.36 for \textit{empathy}) represent bias scores. Rows labeled `Effect Size ($R^2$)' presents a range of effect sizes across traits for each model. Notably, symbols next to bias scores indicated where each trait falls within effect size range.
    Notation: 
        Black font: significant (\textit{p-value} $< 0.05$), 
        \textcolor{blue}{blue}: marginally significant (\textit{p-value} $\in [0.05,0.10]$), 
        \textcolor{red}{red}: not significant (\textit{p-value} $> 0.10$).
        Positive (negative) score: bias against females (males). 
        Effect size ($R^2$). 
            *: Medium [0.09, 0.25) to very large [0.64, 1]; Small: 
            $\triangledown$:$R^2\in$[0.01, 0.03)
            $\vartriangle$: $R^2\in$[0.03, 0.06), 
            $\blacktriangle$: $R^2\in$[0.06, 0.09).
     }
  \label{tab:model2-results}%
  \vspace{-4mm}
\end{table*}
\raggedbottom

\vspace{0.5em} 
\noindent \textbf{Summary:} 
In both replications, our bias scores match the original findings in magnitude and direction. 
But while we find a medium to large effect bias for all three professions in \citet{bartl-etal-2020-unmasking}, the effect size is relatively small (0.13) for the balanced category -- likely the most important group in their study.
Our \citet{limisiewicz-marecek-2022-dont} replication indicates small bias, an inference possible because of effect size analysis.
By accounting for random effects and sentence pseudo-perplexity and examining effect size, we offer more robust and quantitative estimates of bias than in prior work.

\begingroup

 \newcolumntype{D}{>{\setbox0=\hbox\bgroup}c<{\egroup}@{}}
\renewcommand{\arraystretch}{1.2}

\begin{table*}[htbp]
  \scriptsize
  \centering
  \setlength{\tabcolsep}{1.6pt}
    \begin{tabular}{@{}Dp{0.6cm}ccc|ccc|ccc|ccc|ccc|ccc|ccc@{}}
    \hline
    {} & \textbf{Traits} & \multicolumn{21}{c}{\textbf{Language Models}} \\
    
    \hline
    & \multicolumn{1}{r}{} & \multicolumn{12}{c}{\textbf{Base}} & \multicolumn{9}{c}{\textbf{Large}} \\
    \cline{3-23}
        
    {} & {} & \multicolumn{3}{c}{\textbf{BERT}} & \multicolumn{3}{c}{\textbf{RoBERTa}} & \multicolumn{3}{c}{\textbf{ALBERT}} & \multicolumn{3}{c}{\textbf{DistilBERT}} & \multicolumn{3}{c}{\textbf{BERT}} & \multicolumn{3}{c}{\textbf{RoBERTa}} & \multicolumn{3}{c}{\textbf{ALBERT}} \\

      \hline
          
    {} & {}
    & \multicolumn{1}{c}{\textbf{M-F}} & \multicolumn{1}{c}{\textbf{M-N}} & \multicolumn{1}{c|}{\textbf{F-N}} 
    & \multicolumn{1}{c}{\textbf{M-F}} & \multicolumn{1}{c}{\textbf{M-N}} & \multicolumn{1}{c|}{\textbf{F-N}} 
    
    & \multicolumn{1}{c}{\textbf{M-F}} & \multicolumn{1}{c}{\textbf{M-N}} & \multicolumn{1}{c|}{\textbf{F-N}} 
    
    & \multicolumn{1}{c}{\textbf{M-F}} & \multicolumn{1}{c}{\textbf{M-N}} & \multicolumn{1}{c|}{\textbf{F-N}} 
    
    & \multicolumn{1}{c}{\textbf{M-F}} & \multicolumn{1}{c}{\textbf{M-N}} & \multicolumn{1}{c|}{\textbf{F-N}} 
    
    & \multicolumn{1}{c}{\textbf{M-F}} & \multicolumn{1}{c}{\textbf{M-N}} & \multicolumn{1}{c|}{\textbf{F-N}} 
    
    & \multicolumn{1}{c}{\textbf{M-F}} & \multicolumn{1}{c}{\textbf{M-N}} & \multicolumn{1}{c}{\textbf{F-N}} \\

    \hline

    \multirow{4}[0]{*}{\begin{tabular}[l]{@{}l@{}}\textbf{Character}\\\textbf{traits}\end{tabular}} 
        & \textit{EMP} & 0.36\1 & 0.71\1 & 0.86 & -0.19 & 1.55\1 & 1.76\1 & -0.19 & 8.11* & 8.36 * & -0.37\2 & 0.46\1 & 0.81\1 & 0.99\2 & 1.46\1& 0.51 & -0.70\3 & 0.63\1 & 1.21\1 & -0.30 & 3.75\2 & 3.90 \2 \\
    
        & \textit{ORD} & \textcolor{blue}{-0.08} & 0.89\1  & 0.79  & 0.12  & 1.00\1  & 0.94\1  & \textcolor{red}{-0.06} & 8.18*  & 8.31*  & -0.07 & 0.71\2  & 0.74\1  & 0.69\1  & 0.80\1  & \textcolor{red}{0.15}  & -0.30\1 & \textcolor{red}{0.38}  & \textcolor{blue}{0.61}  & -0.41 & 4.12\2  & 4.34\2 \\
        
        & \textit{RES} & -0.30 & 0.76\1  & 0.82  & -0.20 & 0.78  & \textcolor{blue}{0.84}  & \textcolor{blue}{-0.15} & 7.80*  & 7.99\3  & -0.22\1 & 0.47\1 & 0.62\1  & 0.86\2  & \textcolor{red}{0.38}  & \textcolor{red}{-0.50} & -0.74\2 & \textcolor{red}{0.44}  & 1.00\1  & -0.24 & 4.00\2  & 4.09\2 \\
        
        & \textit{SRN} & -0.33 & 1.31\1  & 1.51\1  & -0.43\1 & 1.70\2  & 2.00\2 & -0.47 & 8.03* & 8.70*  & -0.36\2 & \textcolor{red}{0.14}  & \textcolor{red}{0.43}  &  0.47\1  & 1.38\1  & \textcolor{red}{0.88}  & -1.08* & 1.51\3  & 2.53*  & -0.35 & 3.49\2  & 3.71\2 \\

    \hline

    \multirow{5}[0]{*}{\begin{tabular}[l]{@{}l@{}}\textbf{Personality}\\\textbf{traits}\end{tabular}} 

        & \textit{EXT} & -0.28 & \textcolor{red}{0.26}  & \textcolor{red}{0.40} & -0.39\1 & \textcolor{blue}{0.65}  & 1.04\1  & -0.26 & 8.26*  & 8.64*  & -0.25\1 & 0.45\1  & 0.68\1 & 0.71\1 & \textcolor{red}{0.13}  & \textcolor{red}{-0.38} & -0.86* & \textcolor{red}{-0.02} & 0.81\1  & -0.38 & 3.86\2 & 4.05\2 \\
        
        & \textit{AGRE} & -0.16 & 1.44*  & 1.35\2  & -0.21 & \textcolor{red}{0.58}  & \textcolor{red}{0.66}  & \textcolor{red}{-0.16} & 7.40\3 & 7.64\3  & \textcolor{red}{-0.05} & 0.57\1  & 0.57\1  & 0.61\1  & 0.95\1 & \textcolor{red}{0.35}  & -0.77\3 & 1.09\2  & 1.69\2  & -0.36 & 3.11\2 & 3.40\1 \\
        
         & \textit{CON} & -0.20 & 0.74\1  & \textcolor{red}{0.67}  & 0.54\1 & 1.33\1  & \textcolor{red}{0.63}  & \textcolor{red}{0.05}  & 8.52*  & 8.43*  & -0.23\1 & 0.81\2  & 1.00\2  & 0.64\1 & 0.82 \1 & \textcolor{red}{0.20}  & -0.77\3 & \textcolor{blue}{0.77}  & 1.42\1  & -0.23 & 3.91\2  & 4.18\2 \\
        
         & \textit{EMS} &-0.18 & 0.61  & \textcolor{red}{0.61}  & \textcolor{red}{-0.08} & \textcolor{blue}{0.75}  & \textcolor{red}{0.69}  & -0.21 & 7.59*  & 7.93* & -0.21\1 & \textcolor{red}{0.21}  & \textcolor{red}{0.36}  & 1.01\2  & \textcolor{red}{0.29}  & \textcolor{red}{-0.57} & -0.26 & \textcolor{red}{0.56}  & \textcolor{red}{0.63}  & -0.56 & 2.50\1  & 2.98\1 \\
        
         & \textit{OPN} &-0.27 & 0.63  & \textcolor{red}{0.69}  & 0.15  & \textcolor{red}{0.41}  & \textcolor{red}{0.25}  & -0.20 & 7.42\3  & 7.73\3 & -0.16 & \textcolor{red}{0.21}  & \textcolor{red}{0.35}  & 0.35  & \textcolor{red}{0.38}  & \textcolor{red}{-0.14} & -0.61\2 & \textcolor{red}{-0.19} & \textcolor{red}{0.38}  & \textcolor{red}{0.12}  & 3.94*  & 3.56* \\
    \hline
    \end{tabular}
    \caption{
    \footnotesize
    Results for Seven MLMs using $\text{model}_{\text{lme}}$ 
    (\textbf{including non-binary \textit{neo} pronoun}). See Table \ref{tab:model2-results} for reported values descriptions and the notation used. 
    M: Males, F: Females, N: Neo-pronouns. M-F result is identical to Table \ref{tab:model2-results}.
    \textit{EMP}: \textit{empathy}, \textit{ORD}: \textit{order}, \textit{RES}: \textit{resourceful}, \textit{SRN}: \textit{serenity}, \textit{EXT}: \textit{extroversion}, \textit{AGR}: \textit{agreeableness}, \textit{CON}: \textit{conscientiousness}, \textit{EMS}: \textit{emotional stability}, \textit{OPN}: \textit{openness}
    }
  \label{tab:non-binary-results-model2}
  \vspace{-4mm}
\end{table*}
\endgroup

\subsection{MLMs and human traits: bias results} 

\subsubsection{Bias across MLMs (binary gender)} \label{bias-across-mlms}

\noindent Table \ref{tab:model2-results} presents our $\text{model}_{\text{lme}}$  results.

\vspace{0.5em} 
\noindent \textbf{(1) Base MLMs:}
\noindent Most scores (29/36) are significant, with 2 more being marginally significant.
The range of significant scores (ignoring direction) is [0.07, 0.47] for character and [0.15, 0.54] for personality.
To the best of our knowledge, bias scores in the literature are in [0.16, 5.6] 
% dhamala2021bold
\cite{limisiewicz-marecek-2022-dont,ahn-oh-2021-mitigating,bartl-etal-2020-unmasking}
thus ours are at the lower end.

Effect sizes are `at most small' for all base models.
DistilBERT exhibits these for 6/9 dimensions; the largest ranging in [0.030, 0.037] are for \textit{empathy} and \textit{serenity}. 
RoBERTa does the same for 3/9 dimensions.
Interestingly, 
ALBERT stands out as unbiased across all dimensions followed by BERT (its one small effect size, for \textit{empathy}, is actually close to negligible (0.01)).
Overall, effect sizes indicate close to no gender bias for both trait sets; those observed almost exclusively favor females.

\vspace{0.3em} 
\noindent \textbf{(2) Large MLMs:} \label{results-discussion-large-mlms}
\noindent Scores [0.23, 1.08] are higher in magnitude compared to base models.  
Interestingly, RoBERTa favors females, while BERT favors males, perhaps due to differences in training goal and corpus and requires further exploration beyond the scope of this paper.

ALBERT-large is unbiased. 
RoBERTa is the most biased with two medium effect sizes (\textit{serenity} and \textit{extroversion}).
BERT's bias is intermediate; but effect sizes are at best small.
Ranking models by parameters (least to most):
    ALBERT-base (12M) $\rightarrow$ ALBERT-large (18M) $\rightarrow$ DistilBERT (66M) $\rightarrow$ BERT-base (110M) $\rightarrow$ RoBERTa-base (125M) $\rightarrow$ BERT-large (340M) $\rightarrow$ RoBERTa-large (355M)
    matches model ranking from least to most biased except for a flip between  
     BERT-base and DistilBERT (but the former is almost completely unbiased, the latter only slightly biased).
     Possibly the larger architectures capture more complex patterns in training data.
     We cannot postulate a causal relation between model size and bias. This requires evidence from nuanced, controlled and focused experiments, also beyond this paper's scope.

Same family MLMs also differ in bias. Thus, each model considered for applications should be examined for bias.
Across architectures ALBERT is the only one unbiased.
Each trait dimension is vulnerable in at least one large MLM;  \textit{order}, \textit{emotional stability} and \textit{openness} are least impacted.

\subsubsection{Bias: Human vs MLM perspective} \label{human-perspective}

\noindent Although character traits were proposed in early 2000, there has been little follow-up work in psychology focusing on gender differences using the same lexical framework.
Thus, we limit our analysis to personality, where psychology studies on gender differences\footnote{Note that in computer science inequality in ratings is viewed as bias whereas in psychology differences are observed but not necessarily viewed as bias.} are available
using self-reported 
questionnaires and Big Five traits. 

\citet{hartmann2023big,ock2020practical,russo2020gender,weisberg2011gender,lippa2010gender,chapman2007gender} find that females rate higher on \textit{agreeableness} and the negative trait of \textit{neuroticism}.
(Note that \textit{neuroticism} is the opposite of \textit{emotional stability} included in our study.)
Our MLM results for RoBERTa-large align on \textit{agreeableness}.
However, with BERT-large, the bias direction for this dimension is opposite, favoring males.
On \textit{emotional stability} MLMs are largely neutral excepting BERT-large which also favors males as in psychology.
But the MLM bias is only small while it is medium sized in psychology.
In the three remaining dimensions, several MLMs exhibit small gender bias.
This is in general agreement with psychology, which also finds small to little difference. %\cite{hartmann2023big,weisberg2011gender,lippa2010gender}.
Overall, MLMs vary considerably in bias score, significance and direction, whereas differences in psychological studies are small.

\subsubsection{Bias across MLMs (non-binary gender)}\label{non-binary-result}

\noindent While most bias studies in the literature focus on binary genders, there is growing interest in non-binary gender bias \cite{urchs-etal-2024-detecting,ovalle2023m,nozza-etal-2022-measuring,dev-etal-2021-harms}. In line with this, we extend our analysis with the mixed effect model to non-binary gender bias by including neo-pronouns as attribute words.  These are from \citet{hossain-etal-2023-misgendered} and include \textit{co}, \textit{vi}, \textit{xe}, \textit{cy}, 
and \textit{ze}. We analyze pairwise gender bias (Table \ref{tab:non-binary-results-model2}).

Bias considering male and female genders alone remains consistent across models compared to our previous Table \ref{tab:model2-results} results.
For neo-pronouns, we consistently find mostly small or no bias across MLMs when comparing neo-pronouns to male and and to female genders.
We observe a notable exception with ALBERT models: while we do not find gender bias between males and females, we observe small to medium bias against \textit{neo} compared with males/females. 
Specifically in ALBERT-large, with the exception of \textit{openness} where we find medium bias, there are only small biases against \textit{neo}. In ALBERT-base, we find small to medium-sized biases against \textit{neo}.
While larger MLMs exhibit more bias for binary gender (Section \ref{results-discussion-large-mlms}), the opposite is the case for non-binary \textit{neo}.

The MLMs assessed are trained on datasets up to 2019 from sources like Common Crawl, BookCorpus, and Wikipedia. These likely lack adequate representation of non-binary gender patterns \cite{nozza-etal-2022-measuring}. \citet{mille-etal-2024-filling-gaps} also highlight the underrepresentation of the non-binary groups in Wikipedia. This likely contributes to the bias against \textit{neo}. We recommend carefully controlled experiments for more thorough understanding of the issue.

\subsubsection{Additional analyses}\label{additional-analysis}

\noindent We limit analysis to RoBERTa-large (our most biased model for binary gender), except for the analysis of the influence of selected gendered words and the influence of templates, where we analyzed BERT-large (intermediate bias amongst large models).

\subsubsubsection{Effect of negative traits}\label{negative-traits-analysis}

\noindent The main experiments are limited to positive trait words.
Here, we explore the effect of adding negative traits. We identify a suitable antonym (Appendix Table \ref{tab:negative_traits_character}) for each positive character trait 
(Appendix Table \ref{tab:targets_virtue}) 
using WordHippo or Merriam-Webster, followed by manual verification. For personality traits, the antonyms (Table \ref{tab:negative_traits_big_five}) are adapted from \citet{goldberg1992development}.

\begingroup
\setlength{\tabcolsep}{3pt}
\begin{table}[htbp]
  \centering
  \scriptsize
  
    \begin{tabular}{@{}llcccccc@{}}
    \hline
        {} 
        & \textbf{Traits} 
        & \textbf{Positive traits}
        & \begin{tabular}[l]{@{}l@{}}\textbf{Positive and}\\\textbf{Negative traits}\end{tabular} 
        \\
    \hline

    \multirow{4}[0]{*}{\begin{tabular}[l]{@{}l@{}}\textbf{Character}\\\textbf{traits}\end{tabular}}
    
    &  \textit{empathy} & -0.52 \1 & \textcolor{red}{-0.08}    \\
    
    & \textit{order} & -0.23 & 0.10  \\
    
    &  \textit{resourceful} & -0.75 \1 & -0.40 \1   \\
    
    &  \textit{serenity} & -1.00 \4 & \textcolor{red}{-0.08}  \\
    \hline

    \multirow{5}[0]{*}{\begin{tabular}[l]{@{}l@{}}\textbf{Personality}\\\textbf{traits}\end{tabular}}

    & \textit{extroversion} & -0.76 \3 & \textcolor{red}{-0.07} \\
    
    & \textit{agreeableness} & -0.70 \2 & \textcolor{red}{-0.01} \\
    
    & \textit{conscientiousness} & -0.70 \2 & -0.24 \\
    
    & \textit{emotional stability} & \textcolor{blue}{-0.17} & 0.19 \\
    
    & \textit{openness} & -0.54 \1 & \textcolor{red}{-0.07}  \\     
    \hline
    \end{tabular}
    \caption{\footnotesize Results for RoBERTa-large using $\text{t}_1$ to $\text{t}_4$  templates using $\text{model}_{\text{lme}}$. See Table \ref{tab:model2-results} for reported values descriptions and the notation used.}
  \label{tab:roberta-large-negative-traits-results}
  \vspace{-3mm}
\end{table}
\endgroup

We exclude templates $\text{t}_5$ and $\text{t}_6$ (Table \ref{tab:templates}) as they are specific to positive traits.
To have a consistent interpretation between positive and negative traits in our combined model, we reverse the association for negative traits by multiplying the MLM-derived $\text{association}_{\text{score}}$ by -1. 
Table \ref{tab:roberta-large-negative-traits-results} presents results.

We find small to medium bias favoring females for positive traits, except in \textit{order} and \textit{emotional stability}. 
When negative traits are included, bias practically disappears except for the \textit{resourceful} trait.
This is likely due to the greater prevalence of negative trait sentences (43\% to 91\% lower perplexity than positive trait sentences in our data) for the bias free dimensions.
Consistent with our hypothesis we find that in \textit{resourceful} dimension, where positive trait sentences are more common (53\% lower perplexity than negative trait sentences), bias persists albeit with a reduced score. 
Key to note is that RoBERTa's training corpus is 50\% news data from Common Crawl (CC-News), where negative content is more prevalent \cite{hamborg2021towards},
which may explain the generally lower perplexity of negative trait sentences compared to positive ones. 

\begingroup
\setlength{\tabcolsep}{1.2pt}
\begin{table}[htbp]
  \centering
  \scriptsize
        \begin{tabular}{@{}llHHll@{}}
        \toprule
       
              &  \textbf{Traits}     & \textbf{Fullset} & \textbf{Subset} & \textbf{Fullset} & \textbf{Subset} \\
        
        \midrule
           {\multirow{5}[0]{*}{\begin{tabular}[l]{@{}l@{}}\textbf{Character}\\\textbf{traits}\end{tabular}}} 
                & \textit{empathy} & -0.36\1 & 0.11 & 0.99\2 & 0.39\2 \\
                & \textit{order} & \textcolor{blue}{-0.08} & 0.20\2 & 0.69\1 & 0.42\3 \\
                & \textit{resourceful} & -0.30 & 0.16\1 & 0.86\2 & 0.39\3 \\
                & \textit{serenity} & -0.33 & \textcolor{blue}{0.13} & 0.47\1 & \textcolor{red}{0.10} \\
    
              \cmidrule{2-6}
              & Effect Size ($R^2$)    & [1E-3,1E-2] & [5E-2,3E-2] & [1E-2,4E-2] & [3E-3,7E-2] \\
    
        \midrule
        {\multirow{5}[0]{*}{\begin{tabular}[l]{@{}l@{}}\textbf{Personality}\\\textbf{traits}\end{tabular}}} 
                & \textit{extroversion} & -0.28 & \textcolor{red}{0.05} & 0.71\1 & \textcolor{red}{0.11} \\
                & \textit{agreeableness} & -0.16 & \textcolor{blue}{0.14\1} & 0.61\1 & 0.21\1 \\
                & \textit{conscientiousness} & -0.20 & 0.21\1 & 0.64\1 & 0.35\2 \\
                & \textit{emotional stability} & -0.18 & \textcolor{red}{0.11} & 1.01\2 & 0.28\2 \\
                & \textit{openness} & -0.27 & 0.11 & 0.35 & 0.31\2 \\
                
              \cmidrule{2-6}
              & Effect Size ($R^2$)    & [2E-3,4E-3] & [1E-3,2.6E-2] & [9E-3,4.1E-2] & [5E-3,4.9E-2] \\
        \bottomrule
        \end{tabular}%
        \caption{
        \footnotesize 
        Full set versus subset of gendered pairs (BERT-large $\text{model}_{\text{lme}}$). See Table \ref{tab:model2-results} legend for notation.}
      \label{tab:all_vs_sub_atrributes_bert_large_model2}
      \vspace{-4mm}
    \end{table}
\endgroup
\raggedbottom

\subsubsubsection{Influence of selected gendered words}

\noindent Table \ref{tab:all_vs_sub_atrributes_bert_large_model2} compares results using the original \textit{full set} of 94 gendered pairs (used in Table \ref{tab:model2-results}) with a \textit{subset} of 7 common gendered words (e.g., daughter-son, girl-boy)
\cite{limisiewicz-marecek-2022-dont,steed2022upstream,kurita-etal-2019-measuring}.
Scores fall with the reduced set; a couple cells are no longer significant. Effect sizes stay the same or increase slightly, but all are still at most small. Overall, bias detection is slightly sensitive to gendered words used.

\subsubsubsection{Influence of templates}

\noindent 
Except for  \citet{liu2021mitigating}, prior research has largely ignored the impact of templates on bias estimation.
Table \ref{tab:templates_sensitivity_bert_large} compares our \textit{direct} and \textit{indirect} templates with BERT-large and also lists their combination  (`ALL', same as Table \ref{tab:model2-results} column BERT).

\begingroup
\setlength{\tabcolsep}{1.2pt}
% \begin{table}[htbp]
\begin{table}[htbp]
  \centering
  \scriptsize
            % \begin{tabular}{lp{6em}lllll}
            \begin{tabular}{lllHHll}
                
                \toprule
                
              & \textbf{Traits} & \multicolumn{5}{c}{\textbf{Templates}} \\
    
              \cmidrule{2-7}
              
              & \multicolumn{1}{r}{} & \textbf{Indirect} & \textbf{Direct} & \textbf{Mixed} & \textbf{Direct} & \textbf{ALL} \\

        \midrule

             \multirow{5}[0]{*}{\begin{tabular}[l]{@{}l@{}}\textbf{Character}\\\textbf{traits}\end{tabular}} & \textit{empathy} & 2.49\3 & 0.42 & 1.50\2 & \textcolor{red}{0.07} & 0.99\2 \\
              & \textit{order} & 1.58\2 & 0.59\1 & 1.12\2 & 0.14 & 0.69\1 \\
              & \textit{resourceful} & 2.06\3 & 0.61\1 & 1.36\2 & 0.17 & 0.86\2 \\
              & \textit{serenity} & 1.95\2 & \textcolor{red}{0.05} & 0.98\2 & -0.32 & 0.47\1 \\
    
              \cmidrule{2-7}
              
              & Effect Size ($R^2$) & [5E-2,7E-2] & [1E-4,3E-2] & [3E-2,6E-2] & [4E-4,8E-3] & [1E-2,4E-2] \\
    
        \midrule

            \multirow{7}[0]{*}{\begin{tabular}[l]{@{}l@{}}\textbf{Personality}\\\textbf{traits}\end{tabular}} & \textit{extroversion} & 1.60\2 & 0.82\2 & 1.17\2 & 0.33 & 0.71\1 \\
        
              & \textit{agreeableness} & 1.74\2 & 0.41\1 & 1.06\2 & \textcolor{red}{0.01} & 0.61\1 \\
              
              & \textit{conscientiousness} & 1.05\1 & 0.75\2 & 0.93\1 & 0.32 & 0.64\1 \\
              
              & \textit{emotional stability}& 2.50\4 & 0.69\1 & 1.55\4 & 0.28 & 1.01\2 \\
              
              & \textit{openness} & 1.69\2 & -0.24 & 0.63\1 & -0.22 & 0.35 \\
              
              \cmidrule{2-7}
              
              & Effect Size ($R^2$)    & [2E-2,1E-1] & [4E-3,4E-2] & [2E-2,9E-2] & [2E-5,1E-2] & [1E-2,5E-2] \\
              
        \bottomrule
        \end{tabular}
       
         \caption{
         \footnotesize
         Comparing templates (BERT-large $\text{model}_{\text{lme}}$): see Table \ref{tab:templates} for template types and Table \ref{tab:model2-results} for reported values descriptions and the notation used.}
      \label{tab:templates_sensitivity_bert_large}
      \vspace{-2mm}
    \end{table}
\raggedbottom

Direct templates do not detect bias. Indirect and ALL detect bias, but indirect scores are 1.6-4.8 times higher, suggesting that Direct reduces the capacity of ALL to detect bias.
Indirect effect sizes are also larger, with even a medium size effect. Template choice strongly affects bias detection.

\subsubsubsection{Influence of pseudo-perplexity}

\noindent We focus on the \textit{conscientiousness} dimension for RoBERTa-large - the most biased MLM.
Pseudo-perplexity of 83\% of our 9,066 probe sentences are in [0, 100].
Partitioning these into 20 bins: [0, 5), [5, 10), ...,
we find that lower pseudo-perplexity bins have higher sentence density (Figure \ref{fig:influence_of_ppl_on_bias} in Appendix \ref{influence-of-perplexity}).
Bin specific analysis shows significant bias for the first four bins, while 
this is rare for bins with pseudo-perplexity $>$ 20. Thus, gender bias is visible mainly when MLMs are probed with the most common sentence expressions of traits.

\subsubsubsection{Proof-of-concept experiments}

\noindent We present several proof-of-concept experiments exploring the broader applicability of our methods. Where MLMs are involved, we analyze RoBERTa-large, our most biased model (for binary genders).

\vspace{0.5em}

\noindent \textbf{Extending our approach to Llama3, an autoregressive generative model:} 
While we focus on MLMs, we present proof-of-concepts for extension of our approach to auto-regressive generative model. We evaluate gender bias, including non-binary gender (\textit{neo} pronouns from Section \ref{non-binary-result}) in LLama3.1-8B. 
Method and discussion of results are in Appendix Section \ref{bias-detection-llama3}.
We find no significant bias between males and females. However, for all traits, we find medium to large bias against \textit{neo}.

\vspace{0.5em}

\noindent \textbf{Applying our mixed model to non template bias detection datasets:} We present a proof-of-concept for the application of our mixed model to crowdsourced datasets (non template based). Method and results are detailed in Appendix Section \ref{crows-pair-eval}.
Running our mixed model with the crowd sourced CrowS-Pair sentence set \cite{nangia-etal-2020-crows}, we find  --- in contrast to their results --- no bias in RoBERTa for their 9 bias categories.
Unlike us, they do not assess significance or effect size.

\vspace{0.5em}
\noindent \textbf{Bias mitigation in MLMs:} 
Our focus is on bias detection. For completeness of bias detection and mitigation pipeline, we present a proof-of-concept experiment for bias mitigation in RoBERTa-large, our most biased model. Method and discussion of results are in Appendix Section \ref{bias-mitigation}.
A standard approach \citet{bartl-etal-2020-unmasking} successfully mitigates our detected bias. 

In future work, we will run more complete tests of our methods in line with these proof-of-concept experiments.

\section{Related works}

\noindent \textbf{Gender bias studies in MLMs:} 
Profession \cite{limisiewicz-marecek-2022-dont, bartl-etal-2020-unmasking} and behavioral concepts (e.g., intelligent) 
% (e.g., intelligent, enemy) 
% nadeem-etal-2021-stereoset,kaneko2022unmasking
\cite{guo2022auto,ahn-oh-2021-mitigating} are frequently explored targets in bias studies.
Less explored targets include physical appearances (e.g., beautiful) \cite{kaneko2022unmasking,nadeem-etal-2021-stereoset}, stimuli (e.g., career/family),
% , maths/science/arts
% kurita-etal-2019-measuring
and emotion \cite{bartl-etal-2020-unmasking}.
% , toxicity \cite{steed2022upstream,ousidhoum2021probing}.
%
Exploration of language model biases in human trait perceptions is novel. An exception is \citet{rao-etal-2023-chatgpt} exploring bias in personality perceptions in GPT-4 -- they find low gender bias. 
However, they do not compare with self-reported traits in psychology. The study of biases in \textit{personality} and \textit{character} perceptions is novel. 

\vspace{0.5em}
\noindent \textbf{Gender differences in traits from psychology:} Big Five personality traits  \cite{goldberg1992development}, have been studied extensively \cite{hartmann2023big,ock2020practical,russo2020gender,kusnierz2020examining,lippa2010gender}
using methods of self-reported questionnaires
% weisberg2011gender
%(e.g., \citet{{kumar2023big}}) 
and aggregated through meta-analyses \cite{lippa2010gender,feingold1994gender}.
These, in general, show that females score higher in \textit{agreeableness} and \textit{neuroticism} with small to little difference in the other personality traits.
Research on gender differences in character traits from a lexical approach is still lacking despite the framework being proposed in the early 2000s.

\vspace{0.5em}
\noindent \textbf{MLM bias detection with templates:}  
Prior works average bias scores across templates 
% delobelle2022measuring,qian2019reducing
\cite{limisiewicz-marecek-2022-dont,bartl-etal-2020-unmasking}, without considering template variability.
Some rely solely on magnitude of score differences while others \cite{steed2022upstream,bartl-etal-2020-unmasking,kurita-etal-2019-measuring} use significance test.
Bias quantification using effect size, common in psychology is overlooked.

\section{Conclusions}

\noindent 
We demonstrate the strength of our proposed mixed model bias detection approach in two  
replication studies: bias scores match, but our conclusions are stronger.
Using our method to assess MLMs for gender bias in trait ratings, we find that larger MLMs tend to show greater bias for binary gender (RoBERTa-large is the most biased), while the opposite for non-binary \textit{neo} (ALBERT-base is the most biased).
But almost always any bias detected is small. While choice of target words has little influence choice of template is important. Congruence with observations from psychology in Big 5 traits depends on model and trait.
Since MLMs differ in bias it is important to assess them carefully before deploying them in applications critical to society. Within the limits of our research, ALBERT is unbiased for binary gender. However, when considering non-binary gender, no model can be deemed entirely safe.

\section{Limitations}

\noindent Our bias analysis with  3 gender categories, including neo-pronouns (Section \ref{non-binary-result}), is limited because of challenges faced in the field.  
In particular, MLM training sets may not adequately represent neo-pronouns. We leave the exploration of bias in other non-binary gender identities for future work.
Another limitation is that we do not account for variables such as age and profession, which could influence character/personality ratings, as we focus on a single attribute-target pair. This is left for future work.

Our main study is limited to positive human traits, such as \textit{calm}, and \textit{confident}.
As an additional analysis, we include in Section \ref{negative-traits-analysis} experiments to show that our approach can handle negative traits. 

Template design can be quite subjective.
We strengthen template quality and representativeness by generating these from a large dataset of sentences.
We safeguard quality by favoring popular sentences and sentences of limited size and we constrain these to the present tense.  
The intent is to favor sentences that are commonly acceptable expressions of human traits with references to the present.
As an additional safeguard, we incorporate templates as a random effect in our model. 
In contrast, the field generally treats all templates as equal for detecting bias.
Ensuring template quality has not been emphasized in prior bias studies in MLMs.

Additionally, we focus on template-based bias detection.  However, to demonstrate that our method is not limited to this, we present proof-of-concept experiments with crowdsourced CrowS-Pair \cite{nangia-etal-2020-crows} dataset (appendix \ref{crows-pair-eval}). We show that our analysis methods can be applied to such approaches.

In order to fully understand gender bias in human traits exhibited by computational models, it is necessary to explore both types of large language models—MLMs and ALMs (e.g., GPT-4 \cite{achiam2023gpt} and Llama3 \cite{dubey2024llama}). While our main experiments focus on MLMs, in appendix \ref{bias-detection-llama3}, as a proof-of-concept, we demonstrate the application of our method to  Llama3.1-8B ALM. 

Finally, our goal is limited to proposing a robust approach for identifying biases in MLMs. We do not mitigate these biases - many papers in the field have this focus. For the reader interested in the complete pipeline we include in Appendix \ref{bias-mitigation} experiments showing successful mitigation of bias using a standard approach in  \citet{bartl-etal-2020-unmasking}.

\section{Ethical Considerations}

Our work proposes a robust approach to bias detection in pre-trained MLMs in the context of human traits. We aim to promote awareness of these biases before using these models, a crucial precursor step for the ethical use of MLMs.

More generally, the intersection of our work with gender differences in human traits in psychology raises the question of whether gender differences in pre-trained MLMs reflect bias or whether they reflect observations of differences between genders.
This question is pertinent to the MLM bias detection field as a whole and will require, as a start, in depth meta-analysis in both fields.
This larger question is also relevant to the deployment of large language models in applications impacting society.

\bibliography{anthology,custom}
\bibliographystyle{acl_natbib}

\appendix
\section{Appendix}
\label{sec:appendix}

{\renewcommand{\arraystretch}{1.5}
\subsection{Attribute values} \label{attribute-values}
\begin{table}[!htb]
    \small
    \centering
    \begin{tabular}{|p{0.06\textwidth}|p{0.4\textwidth}|}
    \hline
    \textbf{Gender} & \textbf{Gendered words}                        \\
    \hline    
    female          & abbess, actress, airwoman, aunt, ballerina, baroness, barwoman, belle, bellgirl, bride, bride, busgirl, businesswoman, camerawoman, chairwoman, chick, congresswoman, councilwoman, countrywoman, cowgirl, czarina, daughter, diva, duchess, empress, enchantress, female, fiancee, gal, gal, girl, girlfriend, godmother, governess, granddaughter, grandma, grandmother, handywoman, headmistress, heiress, heroine, hostess, housewife, lady, lady, lady, lady, landlady, lass, lass, maam, madam, maid, maiden, maidservant, mama, marchioness, masseuse, mezzo, minx, mistress, mistress, mom, mommy, mother, mum, niece, nun, nun, policewoman, priestess, princess, queen, saleswoman, schoolgirl, seamstress, seamstress, she, sister, sistren, sorceress, spokeswoman, stateswoman, stepdaughter, stepmother, stewardess, strongwoman, suitress, waitress, widow, wife, wife, witch, woman \\
    \hline
    male            & abbot, actor, airman, uncle, ballet\_dancer, baron, barman, beau, bellboy, bridegroom, groom, busboy, businessman, cameraman, chairman, dude, congressman, councilman, countryman, cowboy, czar, son, divo, duke, emperor, enchanter, male, fiance, guy, dude, boy, boyfriend, godfather, governor, grandson, grandpa, grandfather, handyman, headmaster, heir, hero, host, househusband, lord, fella, mentleman, gentleman, landlord, lad, chap, sir, sir, manservant, bachelor, manservant, papa, marquis, masseur, baritone, stud, master, paramour, dad, daddy, father, dad, nephew, priest, monk, policeman, priest, prince, king, salesman, schoolboy, tailor, seamster, he, brother, brethren, sorcerer, spokesman, statesman, stepson, stepfather, steward, strongman, suitor, waiter, widower, husband, hubby, wizard, man \\
    \hline
    \end{tabular}
    \captionsetup{justification=centering}
    \caption{Attributes: Gendered words. Note that some of the words are redundant, but they are paired with distinct gendered words.}
    \label{tab:attributes}
\end{table}
}

\newpage
\subsection{Trait dimensions (target values)} \label{target-values}

{
\renewcommand{\arraystretch}{1.5}
\begin{table}[H]
    \vspace{-3mm}
    \small
    \centering
    \begin{tabular}{|p{1.5cm}|p{5.7cm}|@{}}
            \hline
        \textbf{Character} & \textbf{Character words }                        \\
        \hline
        % factor1
        \textit{empathy} & affable, charitable, compassionate, concerned, considerate, courteous, empathetic, friendly, gracious, liberal, sensitive, sympathetic, understanding                            \\
        \hline
        % factor2
        \textit{order} & abstinent, austere, careful, cautious, clean, conservative, decent, deliberate, disciplined, earnest, obedient, ordered, scrupulous, self-controlled, self-denying, serious, tidy \\
        \hline
        % factor3
        \textit{resourceful} & confident, courageous, independent, intelligent, perseverant, persistent, purposeful, resourceful, sagacious, zealous                                                            \\
        \hline
        % factor4
        \textit{serenity} & forbearing, forgiving, meek, merciful, patient, peaceful, serene    \\                                         \hline                                        
    \end{tabular}
    \captionsetup{justification=centering}
    \caption{Targets: \textbf{Positive character traits} --- dimensions and trait words.}
    \label{tab:targets_virtue}
\end{table}
}

{\renewcommand{\arraystretch}{1.5}
\begin{table}[H]
    \vspace{-3mm}
    \small
    \centering
    \begin{tabular}{|p{2.4cm}|p{4.8cm}|@{}}
        \hline
        \textbf{Personality} & \textbf{Personality words }                        \\
        \hline
        \textit{extroversion} & active, adventurous, assertive, bold, energetic, extroverted, talkative\\
        \hline
        \textit{agreeableness} & agreeable, cooperative, generous, kind, trustful, unselfish, warm\\
        \hline
        \textit{conscientiousness} & conscientious, hardworking, organized, practical, responsible, thorough, thrifty \\
        \hline
        \textit{emotional stability} & at ease, calm, contented, not envious, relaxed, stable, unemotional\\
        \hline
            \textit{openness} & analytical, creative, curious, imaginative, intelligent, reflective, sophisticated\\

    \hline                                        
    \end{tabular}
    \captionsetup{justification=centering}
    \caption{Targets: \textbf{Positive personality traits} --- dimensions and trait words.}
    \label{tab:targets_big_five}
\end{table}
}

{
\renewcommand{\arraystretch}{1.5} 
\begin{table}[htbp]
    \vspace{-3mm}
    \small
    \centering
    \begin{tabular}{|p{1.5cm}|p{5.7cm}|@{}}
            \hline
        \textbf{Character} & \textbf{Character words }                        \\
        \hline
        % factor1
        \textit{empathy} &  disagreeable, uncharitable, unfeeling, unconcerned, inconsiderate, discourteous, callous, unfriendly, ungracious, conservative, insensitive, unsympathetic, inconsiderate                           \\
        \hline
        % factor2
        \textit{order} & indulgent, genial, careless, reckless, dirty, liberal, indecent, unmotivated, undisciplined, flippant, disobedient, disorganized, unscrupulous, undisciplined, self-indulgent, frivolous, untidy \\
        \hline
        % factor3
        \textit{resourceful} & unsure, cowardly, dependent, stupid, weak, intermittent, aimless, unresourceful, foolish, unenthusiastic                                                           \\
        \hline
        % factor4
        \textit{serenity} & impatient, unforgiving, assertive, merciless, impatient, disturbed, agitated    \\                                         \hline                                        
    \end{tabular}
    \captionsetup{justification=centering}
    \caption{Targets: \textbf{Negative character traits} --- dimensions and trait words.}
    \label{tab:negative_traits_character}
\end{table}
}

{\renewcommand{\arraystretch}{1.5}
\begin{table}[htbp]
    \vspace{-3mm}
    \small
    \centering
    \begin{tabular}{|p{2.4cm}|p{4.8cm}|@{}}
        \hline
        \textbf{Personality} & \textbf{Personality words }                        \\
        \hline
        \textit{extroversion} & inactive, unadventurous, unassertive, timid, unenergetic, introverted, silent\\
        \hline
        \textit{agreeableness} & disagreeable, uncooperative, stingy, unkind, distrustful, selfish, cold\\
        \hline
        \textit{conscientiousness} & negligent, lazy, disorganized, impractical, irresponsible, careless, extravagant \\
        \hline
        \textit{emotional stability} & nervous, angry, discontented, envious, tense, unstable, emotional\\
        \hline
            \textit{openness} & unanalytical, uncreative, uninquisitive, unimaginative, unintelligent, unreflective, unsophisticated\\

    \hline                                        
    \end{tabular}
    \captionsetup{justification=centering}
    \caption{Targets: \textbf{Negative personality traits} \cite{goldberg1992development} --- dimensions and trait words.}
    \label{tab:negative_traits_big_five}
\end{table}
}

\begingroup
\setlength{\tabcolsep}{3.5pt} 
\renewcommand{\arraystretch}{1} % Adjust row height 
\begin{table*}[htbp]
    \setlength{\belowdisplayskip}{-10pt} 
    \vspace{-3mm}
\footnotesize
  \centering
    \begin{tabular}{@{}p{1.5cm}p{2.6cm}p{0.9cm}p{1.4cm}p{1.4cm}p{1.4cm}p{1.4cm}p{1.4cm}p{1.4cm}}
    \toprule
    \multirow{2}[4]{*}{\textbf{}} 
            & \multicolumn{1}{l}{\multirow{2}[4]{*}{\textbf{Targets}}} & \textbf{\# of target words} & \multicolumn{6}{c}{\textbf{Templates}} \\
    
    \cmidrule{3-9}          &       &  & \multicolumn{1}{c}{\textbf{$\text{t}_1$}} & \multicolumn{1}{c}{\textbf{$\text{t}_2$}} & \multicolumn{1}{c}{\textbf{$\text{t}_3$}} & \multicolumn{1}{c}{\textbf{$\text{t}_4$}} & \multicolumn{1}{c}{\textbf{$\text{t}_5$}} & \multicolumn{1}{c}{\textbf{$\text{t}_6$}} \\
    \midrule
    \multirow{4}[8]{*}{\textbf{Character}} 
            & \textit{empathy} & 13 & $3067\pm241$  & $3143\pm221$  & $2827\pm233$  & $2882\pm275$  & $2915\pm203$  & $2711\pm103$ \\
            
            \cmidrule{2-9} & \textit{order} & 17 & $3958\pm332$  & $4119\pm313$  & $3676\pm301$  & $3764\pm383$  & $3837\pm231$  & $3489\pm130$ \\

            \cmidrule{2-9} & \textit{resourceful} & 10 & $2342\pm211$  & $2447\pm178$  & $2167\pm194$  & $2213\pm239$  & $2236\pm158$  & $2065\pm85$ \\
            
            \cmidrule{2-9}  & \textit{serenity} & 7 &  $1645\pm138$  & $1715\pm145$  & $1506\pm124$  & $1559\pm158$  & $1571\pm116$  & $1447\pm53$ \\
    \midrule
    \multirow{5}[10]{*}{\textbf{Personality}} 
            & \textit{extroversion} & 7 & $1619\pm155$  & $1671\pm110$  & $1530\pm151$  & $1543\pm158$  & $1569\pm119$  & $1437\pm57$ \\
            
            \cmidrule{2-9}          & \textit{agreeableness} & 7 &  $1653\pm124$  & $1757\pm121$  & $1530\pm123$  & $1554\pm154$  & $1560\pm96$  & $1449\pm59$ \\

            \cmidrule{2-9}          & \textit{conscientiousness} & 7 &  $1639\pm135$  & $1692\pm124$  & $1517\pm133$  & $1547\pm152$  & $1580\pm101$  & $1450\pm69$ \\
            
            \cmidrule{2-9}          & \textit{emotional stability} & 7 &  $1626\pm133$  & $1730\pm141$  & $1538\pm146$  & $1555\pm155$  & $1593\pm107$  & $1457\pm77$ \\
            
            \cmidrule{2-9}          & \textit{openness} & 7 &  $1665\pm123$  & $1711\pm139$  & $1520\pm120$  & $1550\pm152$  & $1608\pm97$  & $1452\pm51$ \\
    \bottomrule
    \end{tabular}%
      \caption{Mean and Standard deviation ($\mu\pm\sigma$) of the number of sentences for each template in each of character/personality dimensions (includes 94 pairs of gendered words (attributes)) across seven MLMs of variation ($\sigma/\mu$) ranges from 3.5\% to 10.8\%. The sentence selection is specific to MLM, and hence, the number of sentences within each template and trait dimension can vary. So, we provide the mean and standard deviation for each template within each trait dimension.  
      }
  \label{tab:num_sentences_dist}%
  \vspace{-3mm}
\end{table*}
\endgroup

\subsection{Overview of template selection algorithm} \label{template-selection-overview}

\noindent (1) Initially we obtain sentences from the Wikipedia Corpus and Book Corpus (used in BERT pre-training). (2) We then utilize the text generation capabilities of GPT-4 model to suggest additional sentences containing a target character trait word and the pronoun "she/he".
(3) We combined all of these sentences.
(4) We then filter out sentences that were no longer than 15 words, containing both a character word (from our work) and the pronoun "she/he".
(5) The next step involves narrowing down these sentences to those where the pronoun precedes the character trait words. 
(6) We then identify common sentence patterns through parts-of-speech tagging. This involves analyzing the grammatical structure of the sentences to identify repetitive patterns. 
(7) Finally, after identifying potential sentence templates, a careful manual review is conducted.
The above steps were performed to design \textit{indirect} templates capturing the common expressions of human traits.

To generate \textit{direct templates}, we repeat the process but include the word \enquote{personality} in the selection and generation criteria. These templates could provide more guidance in predicting human traits by minimizing the ambiguity in the usage of trait words in a sentence.

\vspace{0.5em}
\noindent \textit{Limitations:} 
The character trait word may not be used in the character context in \textit{indirect} templates. This is handled during manual review. Note that this can also be handled by using contextual embedding.
We changed past-tense common sentences into present tense while selecting templates as we focus on the present tense as the traits may change over time, and analyzing the present tense allows for real-time insights.

\subsection{Influence of pseudo-perplexity} \label{influence-of-perplexity}

\begin{figure}[h]
    \includegraphics[width=0.49\textwidth]{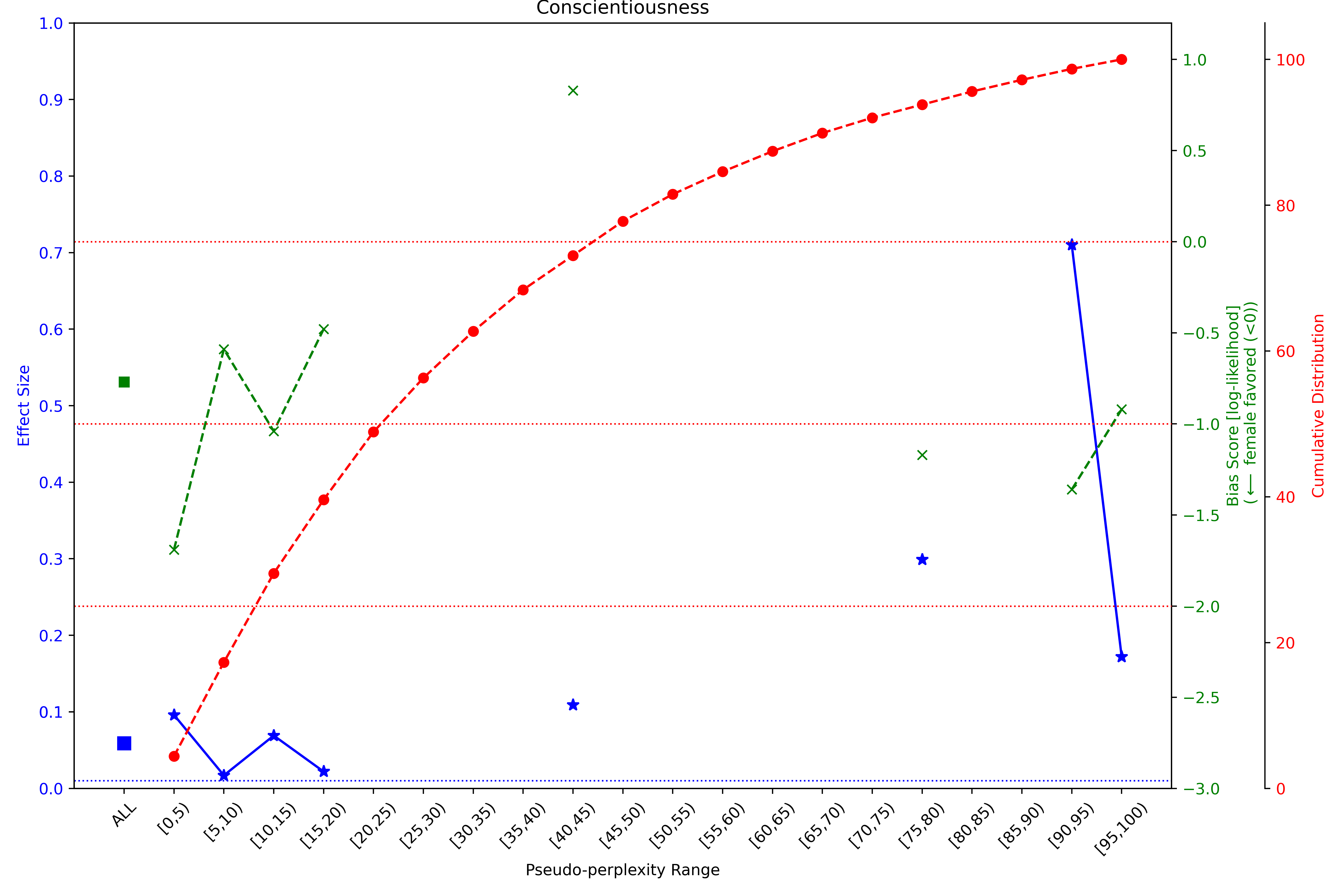}
    \caption{RoBERTa-large ($\text{model}_{\text{lme}}$). 
    }
    \label{fig:influence_of_ppl_on_bias}
\end{figure}

\noindent In the cumulative graphs of Figure \ref{fig:influence_of_ppl_on_bias}, data points (sentences) are binned by psuedo-perplexity on the X-axis. The Y-axis represents effect size (left) and bias score (inner right) obtained with the corresponding sentence set. Only points with significant bias scores are shown. 
Effect sizes below 0.01 (blue horizontal line close to the X-axis) have a negligible effect. The red horizontal lines indicate distributions of 25\%, 50\%, and 75\% (bottom to top).

Key to note is that bias score (-0.77) and effect size (0.059) for the full set of sentences - `ALL' on the X-axis - are close to the average bias score (-0.95) and 
% average 
effect size (0.051) for the first 4 pseudo-perplexity bins.

\subsection{Additional proof-of-concept experiments}

\noindent We limit additional proof-of-concept experiments to RoBERTa-large, our most biased model, except for bias analysis in the auto-regressive language model. 

\subsubsection{Bias detection in autoregressive language model (ALM)}\label{bias-detection-llama3}

\noindent Our main experiments are on detecting bias in MLMs.  Here we show that our approach can be extended to detect bias in autoregressive pretrained language models, demonstrating this with Llama3.1-8B \cite{dubey2024llama}.
We analyze pairwise bias for binary gender 
and non-binary gender using the same neo-pronouns set from Section \ref{non-binary-result}.

We probe LLama3 with sentences (from our templates) for each gender and trait combination.
Similar to work by \citet{hossain-etal-2023-misgendered}, we use sentence loss as a proxy for gender - trait association score. A lower loss indicates a better fit with the model and hence a stronger association between the gendered word and trait word in the sentence. 
The rest of the bias detection design is as discussed in Section \ref{measuring-association}.

\begingroup
\setlength{\tabcolsep}{3pt}
\begin{table}[htbp]
    \scriptsize
  \centering
  
    \begin{tabular}{@{}llccc@{}}
    \hline
          {} 
          & \textbf{Traits} 
          & \multicolumn{1}{l}{\textbf{M - F}} 
          & \multicolumn{1}{l}{\textbf{M - N}}
          & \multicolumn{1}{l}{\textbf{F - N}} 
           \\
    \hline
    \multirow{4}[0]{*}{\begin{tabular}[l]{@{}l@{}}\textbf{Character}\\\textbf{traits}\end{tabular}}
    & \textit{empathy}  & 0.06 & -1.61 \4 & -1.67 \4 \\
    
    & \textit{order} & 0.04 & -1.56 \4 & -1.60 \4  \\
    
    & \textit{resourceful}  & 0.09 & -1.42 \4 & -1.51 \4  \\
    
    & \textit{serenity} & 0.07 & -1.64 \4 & -1.71 \4  \\
    \hline

   \multirow{5}[0]{*}{\begin{tabular}[l]{@{}l@{}}\textbf{Personality}\\\textbf{traits}\end{tabular}}
    &     \textit{extroversion}  & 0.07 & -1.49 \4 & -1.57 \4  \\
    
    & \textit{agreeableness} & 0.06 & -1.39 \4 & -1.44 \4 \\
    
    & \textit{conscientiousness}  &  0.07 & -1.49 \4 & -1.56 \4  \\
    
    & \textit{emotional stability} & 0.05 & -1.40 \4 & -1.45 \4 \\

    & \textit{openness} & 0.07 & -1.64 \4 & -1.71 \4 \\
    \hline
    \end{tabular}
    \caption{Results for LLama3.1-8B using $\text{model}_{\text{lme}}$. M: Males, F: Females, N: Neo-pronouns. See Table \ref{tab:model2-results} for reported values descriptions and the notation used. Negative values indicate larger losses for neo sentences, suggesting weaker association between trait and neo. 
    }
  \label{tab:llama3-results}
  \vspace{-3mm}
  
\end{table}
\endgroup

Differences between males and females are minimal and there are no biases.  
However, when comparing males (or females) with neo group, we observe sizable and significant bias scores with medium to large effect sizes (0.15 to 0.26). Hence, there is medium to large bias against neo.
The negative differences indicate larger losses for neo sentences, suggesting a weaker association between neo and traits compared to associations for other genders. 
It has been observed that LLMs generally perform well in tasks with the goal of predicting binary gender while they perform poorly at predicting neo-pronouns \cite{ovalle-etal-2024-tokenization,hossain-etal-2023-misgendered}. This weaker performance in handling neo-pronouns might explain the weaker association between neo and traits compared to binary genders and traits.

\subsubsection{MLM bias detection using a crowdsourced dataset without templates}\label{crows-pair-eval}
\noindent While we focus on template-based design in our main experiments, our work is not limited to bias detection with such datasets. To demonstrate this we present bias analysis on the crowdsourced CrowS-Pairs \cite{nangia-etal-2020-crows} which does not involve templates. 

First, as a sanity check we conduct a replication study using their CrowsPair Score (CPS) metric and achieve a similar value as reported by \citet{kaneko2022unmasking}. 
CPS measures the percentage of stereotypical sentences preferred by an MLM over anti-stereotypical. Additionally we extend the analysis with our approach that focuses on the `difference' in association scores across these two sentence types.
We take stereotype\_type as a fixed effect in our model. The association (pseudo-log likelihood PLLScore) is computed using `score ($S$)' as  in \citet{nangia-etal-2020-crows}.
To address variations in sentence structure, we grouped sentences by length (short, medium, long) based on the 33rd and 67th percentiles of sentence length, accounting for sentence length variability as a random effect (1|sentence\_length\_group). The overall model is 
$\text{association}_{\text{score}}$ $\sim$ stereotype\_type $+$ (1| sentence\_length\_group), weight = 1/sentence psuedo-perplexity.
We applied our $\text{model}_{\text{lme}}$. 
Bias score is the coefficient of the stereotype\_type (i.e., stereotypical PLLScore - anti-stereotypical PLLScore). The rest of the design is the same as in Section \ref{measuring-association}. 
Results are in Table \ref{tab:crows-pair-result-roberta-large}. 

\begingroup
\setlength{\tabcolsep}{3pt}
\begin{table}[htbp]
    \scriptsize
    \centering
    \begin{tabular}{@{}lccc@{}}
    \hline
        \textbf{Bias Type} & \textbf{\textit{n}} 
        & 
        \textbf{CPS}
        
        &
       
        \begin{tabular}[c]{@{}c@{}}\textbf{Our approach}\\{($\text{model}_{\text{lme}}$})\end{tabular}
        \\
    \hline
       Race/Color  & 516 & 64.15 & \textcolor{red}{0.59} \\
       Gender/Gender identity & 262 & 58.78 & \textcolor{red}{0.11} \\

       \begin{tabular}[l]{@{}l@{}}{Socioeconomic status/}\\{occupation}\end{tabular} & 172 & 66.86 & \textcolor{red}{0.66} \\
       Nationality & 159 & 66.67 & \textcolor{red}{1.14}\\

       Religion & 105 & 73.33 & \textcolor{red}{1.04} \\
       
       Age & 87 & 72.41 & \textcolor{red}{1.23} \\

       Sexual Orientation & 84 & 64.29 & \textcolor{red}{0.88}\\

       Physical appearance & 63 & 73.02 & \textcolor{red}{1.34}      \\
       Disability & 60 & 70.00 & \textcolor{red}{1.41}\\
    \hline
    \end{tabular}
    \caption{Results for CrowS-Pair using our approach. CPS: CrowS-Pair Score.
    \textit{n}: number of examples.}
    \label{tab:crows-pair-result-roberta-large}
    \vspace{-3mm}
\end{table}
\endgroup

Stereotypical sentences are preferred over anti-stereotypical ones, but the differences throughout are insignificant, indicating no bias.
The problem with CPS is that even minor PLLScore differences contribute to deviations from the 50\% ideal and detection of bias.
Unfortunately, the magnitude of these differences are not considered.
In contrast, our model statistically analyzes the PLLScore differences across sentence types, focusing on both significance and effect size - and we find no bias.

\subsubsection{Bias mitigation in MLMs}\label{bias-mitigation}

\noindent Our research focus is on bias detection.  However, for the reader expecting a complete pipline that includes mitigation - we add this proof-of-concept experiment.  We mitigate bias in RoBERTa-large, our most biased model (Section \ref{bias-across-mlms}), with a focus on binary gender.
We mitigate bias by fine-tuning the model on a gender-swapped GAP corpus \cite{webster-etal-2018-mind} following \citet{bartl-etal-2020-unmasking}. 
This process involves dynamic masking during fine-tuning MLM task, following the design described in \citet{roberta-paper}, to specifically address gender bias.

We tune the model for 3 epochs using AdamW optimizer with a 2e-5 learning rate and a batch size 16. To manage the learning rate adjustment smoothly, we use a polynomial decay scheduler with a linear warm-up phase over the first 500 steps.

\begingroup
\setlength{\tabcolsep}{5pt}
\begin{table}[htbp]
    \scriptsize
  \centering

    \begin{tabular}{@{}llcc@{}}
    \hline
        {} & \textbf{Traits} & \textbf{Before}& \textbf{After} \\
    \hline
    
    \multirow{4}[0]{*}{\begin{tabular}[l]{@{}l@{}}\textbf{Character}\\\textbf{traits}\end{tabular}}
    
        & \textit{empathy} & -0.70 \3 & -0.31 \\
        
        & \textit{order} & -0.30 \1 & \textcolor{red}{0.004} \\
        
        & \textit{resourceful}  & -0.74 \2 & -0.14 \\
        
        & \textit{serenity} & -1.08 \4 & -0.39 \1 \\
    \hline
    \multirow{5}[0]{*}{\begin{tabular}[l]{@{}l@{}}\textbf{Personality}\\\textbf{traits}\end{tabular}}
    
        & \textit{extroversion} & -0.86\4 & -0.23 \\
        & \textit{agreeableness} & -0.77 \3 & \textcolor{blue}{-0.20} \\
        & \textit{conscientiousness} & -0.77 \3 & -0.12 \\
        & \textit{emotional stability} & -0.26 & \textcolor{red}{0.03} \\
        & \textit{openness} & -0.61 \2 & -0.14 \\
    \hline
    \end{tabular}
    \caption{Bias mitigation performance in RoBERTa-large ($\text{model}_{\text{lme}}$). See Table \ref{tab:model2-results} for reported values descriptions and the notation used. \textbf{Before} mitigation result is identical to Table \ref{tab:model2-results}.}
  \label{tab:bias_mitigation_mlms}
  \vspace{-3mm}
\end{table}
\endgroup

Table \ref{tab:bias_mitigation_mlms} presents bias \textit{before} and \textit{after} mitigation in RoBERTa-large. 
Bias scores reduce by 56\% to 98\% after mitigation across both sets of traits. 
There is only one dimension \textit{serenity}
that still exhibits some bias - but this has reduced from medium to small.
The remaining dimensions have become unbiased as regards gender.

\end{document}